\documentclass[a4paper, 10pt, conference]{ieeeconf}      
\usepackage{FG2021}

\FGfinalcopy 

\IEEEoverridecommandlockouts                              
\def\FGPaperID{274} 

\usepackage{color, colortbl}
\usepackage{graphicx}
\definecolor{Best}{rgb}{0.976,0.847,0.713}
\definecolor{UM}{rgb}{0.7137,0.85,0.976}
\usepackage{amsmath} 
\usepackage{dblfloatfix}
\usepackage{amssymb}
\usepackage[accsupp]{axessibility} %

\title{\LARGE \bf
Affect-Aware Deep Belief Network Representations\protect\\for Multimodal Unsupervised Deception Detection
}

\author{\parbox{16cm}{\centering
    {\large Leena Mathur and Maja J Matarić}\\
    {\normalsize
    Department of Computer Science, Mathematics and Computer Science, University of Southern California\\}}
}

\begin{document}

%
%
%




\IEEEoverridecommandlockouts\pubid{\makebox[\columnwidth]{978-1-6654-3176-7/21/\$31.00~\copyright{}2021 IEEE \hfill}
\hspace{\columnsep}\makebox[\columnwidth]{ }}

\ifFGfinal
\thispagestyle{empty}
\pagestyle{empty}
\else
\author{Anonymous FG2021 submission\\ Paper ID \FGPaperID \\}
\pagestyle{plain}
\fi
\maketitle

\begin{abstract}
Automated systems that detect the social behavior of deception can enhance human well-being across medical, social work, and legal domains. Labeled datasets to train supervised deception detection models can rarely be collected for real-world, high-stakes contexts. To address this challenge, we propose the first unsupervised approach for detecting real-world, high-stakes deception in videos without requiring labels. This paper presents our novel approach for affect-aware unsupervised Deep Belief Networks (DBN) to learn discriminative representations of deceptive and truthful behavior. Drawing on psychology theories that link affect and deception, we experimented with unimodal and multimodal DBN-based approaches trained on facial valence, facial arousal, audio, and visual features. In addition to using facial affect as a \textit{feature} on which DBN models are trained, we also introduce a DBN training procedure that uses facial affect as an \textit{aligner} of audio-visual representations. We conducted classification experiments with unsupervised Gaussian Mixture Model clustering to evaluate our approaches. Our best unsupervised approach (trained on facial valence and visual features) achieved an AUC of 80\%, outperforming human ability and performing comparably to fully-supervised models. Our results motivate future work on unsupervised, affect-aware computational approaches for detecting deception and other social behaviors in the wild.
\end{abstract}

\section{INTRODUCTION}
Recent developments in representation learning, affective computing, and social signal processing are advancing the creation of automated systems that can perceive human behaviors \cite{10.1145/1459359.1459573, mmml, picard2000affective}. The social behavior of \textit{deception} involves the intentional transmission of false or misleading information during social interactions \cite{ekmanhs}, characterized as either high-stakes or low-stakes. Deceivers in high-stakes contexts face substantial consequences if their deception is discovered, in contrast to those in low-stakes contexts. Human deception detection ability across diverse contexts has been determined as close to chance level \cite{Bond2006AccuracyOD}, motivating the creation of computational systems to help humans with this task. Healthcare providers, social workers, and legal groups can benefit from automated deception detection systems for applications that enhance human well-being (e.g., legal teams assessing courtroom testimonies of children who may be coerced to lie \cite{durante}; social workers and therapists recognizing when people are concealing abusive experiences \cite{vrij2006police}). 

A key challenge in automated deception detection is the \textit{scarcity} of labeled, real-world high-stakes deception data available for training models \cite{porter_brinke}. These data cannot be ethically obtained from lab experiments, because simulating realistic high-stakes scenarios with lab participants requires the use of threats to impose substantial consequences on deceivers. It is also difficult to collect large amounts of labeled deception data with verifiable ground truth from real-world situations. It is worth noting that scarcity of labeled training data also poses a challenge for scientists and engineers seeking to model other affective and behavioral states; our paper contributes to addressing this broader problem. 

We address the scarcity of labeled high-stakes deception data by proposing the first fully-unsupervised approach for detecting high-stakes deception in-the-wild, without requiring labeled data. We experimented with unsupervised \textit{energy-based approaches} using Deep Belief Networks (DBN) \cite{srivastava2012multimodal, salakhutdinov2009deep}, to learn discriminative representations of truthful and deceptive behavior when trained on multimodal behavioral features. DBN-based approaches have shown potential for learning useful representations for human-centered tasks \cite{6638346, 6639140, 10.1109/CVPR.2014.299, 10.1145/2647868.2654969}. To the best of our knowledge, our paper presents the first unsupervised DBN-based approaches for learning representations of a social behavior. 

Prior approaches for deception detection in videos have primarily used features from visual, vocal, verbal, and physiological modalities \cite{10.1145/3107990.3108005}. 
There is limited research that includes affect as a modality for deception detection \cite{10.1145/3382507.3418864, Amiriparian+2016}. Affect refers to neurophysiological states that function as components of emotions and moods. Inspired by psychology research on relationships between affect and social behavior \cite{forgas1995mood}, specifically relationships between affect and deception \cite{ekman1969nonverbal, zuckerman1981verbal}, we chose to include affect in our models. Affective states are typically computationally represented along two dimensions: how pleasant or unpleasant a state is (valence) and how passive or active a state is (arousal) \cite{russell1980circumplex}. Since the face is a key channel through which affect is expressed \cite{App2011NonverbalCU}, our research was scoped to include \textit{facial valence} and \textit{facial arousal} as affect features. In this paper, we report on our experiments with unimodal and multimodal DBN models trained on features from facial valence, facial arousal, audio, and visual modalities, to identify effective modeling approaches. In addition to using facial affect as a \textit{feature} on which our DBN models are trained, we also introduced a novel DBN training procedure to use facial affect as an \textit{aligner} of audio-visual representations. In this paper, we refer to models that use affect as either a feature or an aligner as  \textit{affect-aware} models. Our research is driven by the questions: \textit{To what extent will unsupervised DBN models be effective in learning discriminative representations of deceptive and truthful behavior from videos in-the-wild? How does facial affect contribute to the quality of DBN-based representations when used as a feature or aligner?} 

To evaluate the usefulness of our DBN-based approaches for learning discriminative representations of deceptive and truthful behavior, we conducted classification experiments with unsupervised Gaussian Mixture Models (GMM). Our best unsupervised approach (trained on facial valence and visual features) achieved an AUC of 80\%, outperforming human ability and performing comparably to fully-supervised models. The highest-performing representations for each DBN architecture included facial affect as either a \textit{feature} or an \textit{aligner}, supporting the inclusion of affect in models for learning discriminative representations of deceptive and truthful behavior. Our findings motivate future work on unsupervised, affect-aware machine learning approaches for detecting high-stakes deception and other social behaviors.

This work makes the following contributions:
\begin{itemize}
\item The first unimodal and multimodal DBN-based approaches to learn discriminative representations of a social behavior (deception in high-stakes situations).  
\item A novel unsupervised approach for high-stakes deception detection without requiring labeled data, serving as a proof-of-concept for future research on unsupervised models of deception, and other social behaviors.
\item Introduction and analysis of effective modeling approaches for affect-aware unsupervised deception detection, using facial affect as a feature and as an aligner.
\end{itemize}

\section{RELATED WORKS}
\subsection{Multimodal Deception Detection}
Prior approaches for high-stakes deception detection in videos have largely relied on \textit{supervised} machine learning models that exploit discriminative patterns in visual, vocal, verbal, and physiological cues to distinguish deceptive and truthful communication \cite{10.1145/3107990.3108005, 10.1145/2818346.2820758,10.1145/3349801.3349806, 7836768, RillGarca2019HighLevelFF, 8953413, DBLP:conf/aaai/WuSDS18, 10.1145/3382507.3418864}. Recent approaches have found affect to be a promising modality in deception detection \cite{10.1145/3382507.3418864, Amiriparian+2016}. Approaches that include affect as a \textit{feature} draw on psychology theories that deceivers in high-stakes situations will exhibit affective states with lower valence and higher arousal,
compared to truthful speakers, expressed through patterns in nonverbal cues \cite{ekman1969nonverbal, zuckerman1981verbal}. We build on this research to treat affect not only as a \textit{feature} on which models are trained, but also as an \textit{aligner} of audio-visual representations of deception. Social psychology theories have proposed infusing affect in models of social behavior \cite{forgas1995mood}; our research is driven by these insights. Our paper presents the first attempt at considering affect as an \textit{aligner} of audio-visual feature representations for detecting a social behavior, including deception. 

Existing high-stakes deception detection models have been trained on the only publicly-available, real-world high-stakes deception dataset that is the current benchmark: 121 videos of people speaking in courtroom situations (60 truthful and 61 deceptive videos, $\sim$28 seconds per video) \cite{10.1145/2818346.2820758}. Benchmark accuracies for this dataset \cite{10.1145/2818346.2820758} range from 60-75\% with decision-trees trained on manually-annotated hand gestures, head movements, and facial features, in addition to automatically-extracted transcript n-grams. Prior approaches have developed supervised deep classification models \cite{8953413}; however use of deep classifiers has been considered inadvisable, due to the dataset's small size \cite{DBLP:conf/aaai/WuSDS18}. In this paper, we use this benchmark dataset and compare our unsupervised GMM classifiers to approaches that use automated supervised classical machine learning models trained on interpretable features from human communication, with this dataset (\textbf{Table I}). Prior classifiers have included Decision Trees (DT), Support Vector Machines (SVM), Adaptive Boosting (AB), and Random Forests (RF). 

\begin{table}[h]
\centering
    \caption{Highest-performing automated deception detection approaches using supervised classical machine learning models trained on the benchmark dataset \cite{10.1145/2818346.2820758}.}
    \begin{tabular}{l|l|c|c}
        \textbf{Modality}&\textbf{Model}&\textbf{AUC}&\textbf{ACC}\\
        \hline
         Visual, Verbal \cite{10.1145/2818346.2820758} (benchmark) & DT & --- & 75\%\\
         \hline
          Visual \cite{10.1145/3349801.3349806} & SVM &--- & 77\%\\
          \hline
         Visual, Vocal  \cite{RillGarca2019HighLevelFF} & SVM & 70\%& ---\\
          \hline
         Visual, Vocal, Verbal \cite{DBLP:conf/aaai/WuSDS18}&SVM &88\% & ---\\
         \hline
         Visual, Vocal, Verbal \cite{7836768}& SVM &--- & 79\%\\
          \hline
          Affect, Visual, Vocal \cite{10.1145/3382507.3418864}& AB  & 91\% & 84\%\\
          \hline
         Visual, Vocal, Verbal \cite{9165161} & RF & --- & 68\%\\
        \hline
    \end{tabular}
    \label{tab:my_label}
\end{table}

To address the scarcity of labeled deception data for training models, researchers have developed supervised models that are more robust to small numbers of samples by discriminative feature subspaces \cite{8621909}, adversarial meta-learning \cite{8953413}, and data augmentation \cite{8953413}. These approaches are constrained by dependency on labeled data to train models. Recently, \cite{9413550} has demonstrated the use of domain adaptation to develop models trained on labeled low-stakes deception datasets that detect high-stakes deception; this, too, is constrained by dependence on labeled low-stakes deception data to train models. To the best of our knowledge, we introduce the first unsupervised approaches that detect high-stakes deception \textit{without requiring any labels for training models}.
\subsection{Multimodal Deep Belief Networks}
Deep learning models have demonstrated potential to learn useful representations of visual, verbal, and vocal communication \cite{mmml, bengio2013representation}. Models based on DBNs have emerged as promising approaches for multimodal representation learning. DBNs are probabilistic graphical models that are composed of layers of stochastic, latent variables \cite{salakhutdinov2009deep}. As generative models \cite{salakhutdinov2015learning}, DBNs learn complex, non-linear dependencies across high-dimensional input feature spaces \textit{without requiring labeled data}. DBNs can, therefore, be trained with fully-unsupervised algorithms. DBNs are formed by stacking Restricted Boltzmann Machines (RBMs), which are shallow 2-layer undirected probabilistic graphical models that contain a layer of visible units, a layer of hidden units, a weight matrix of connections between each visible unit and hidden unit, and biases for visible and hidden layers. RBMs assign an \textit{energy} to each configuration of visible and hidden states and are trained to minimize this energy. In a standard DBN, the output of each RBM serves as the input for the next layer's RBM; DBNs can be trained layer-wise with an efficient greedy algorithm \cite{hinton2006fast, salakhutdinov2012efficient}.  

Multimodal DBNs are trained to learn joint representations across modalities that capture complex nonlinear dependencies in input data \cite{srivastava2012multimodal}. Prior research has used multimodal DBNs to combine audio-visual information across modalities through early fusion and late fusion. In \textit{early fusion}, DBNs are trained on inputs concatenated across modalities. In \textit{late fusion}, separate, unimodal DBN layers are trained to learn modality-specific representations before combining these to form a joint input for a multimodal DBN \cite{10.1109/CVPR.2014.299, 6638346, 6639140, 10.1145/2647868.2654969}. These DBN approaches have shown the ability to learn useful representations for human-centered modeling tasks, including emotion recognition \cite{6638346}, pose estimation \cite {10.1109/CVPR.2014.299}, and gesture recognition \cite{10.1145/2647868.2654969}. This paper presents the first multimodal DBN approaches for learning representations of a social behavior, specifically high-stakes deception. 

A key challenge in multimodal representation learning is developing models that \textit{align} information across modalities to learn richer representations \cite{mmml}. The Kabsch algorithm provides a fundamental approach for aligning representations \cite{kabsch1976solution}; it estimates the best rotation matrix to align representations by minimizing their root mean squared deviation. Approaches based on the Kabsch algorithm have been developed to align representations for tasks across diverse fields, including graphics \cite{taylor2012vitruvian}, speech \cite{Kheder2016}, and natural language processing \cite{glavas-vulic-2020-non}. We contribute the first approach for affect-aligned DBN representation learning, adapting the Kabsch algorithm to learn audio, visual, and audio-visual DBN representations of a social behavior (deception) that are aligned with affect representations. 

\section{METHODOLOGY}
\begin{figure*}[b]
   \centering
   \includegraphics[width=0.85\textwidth, height=180pt]{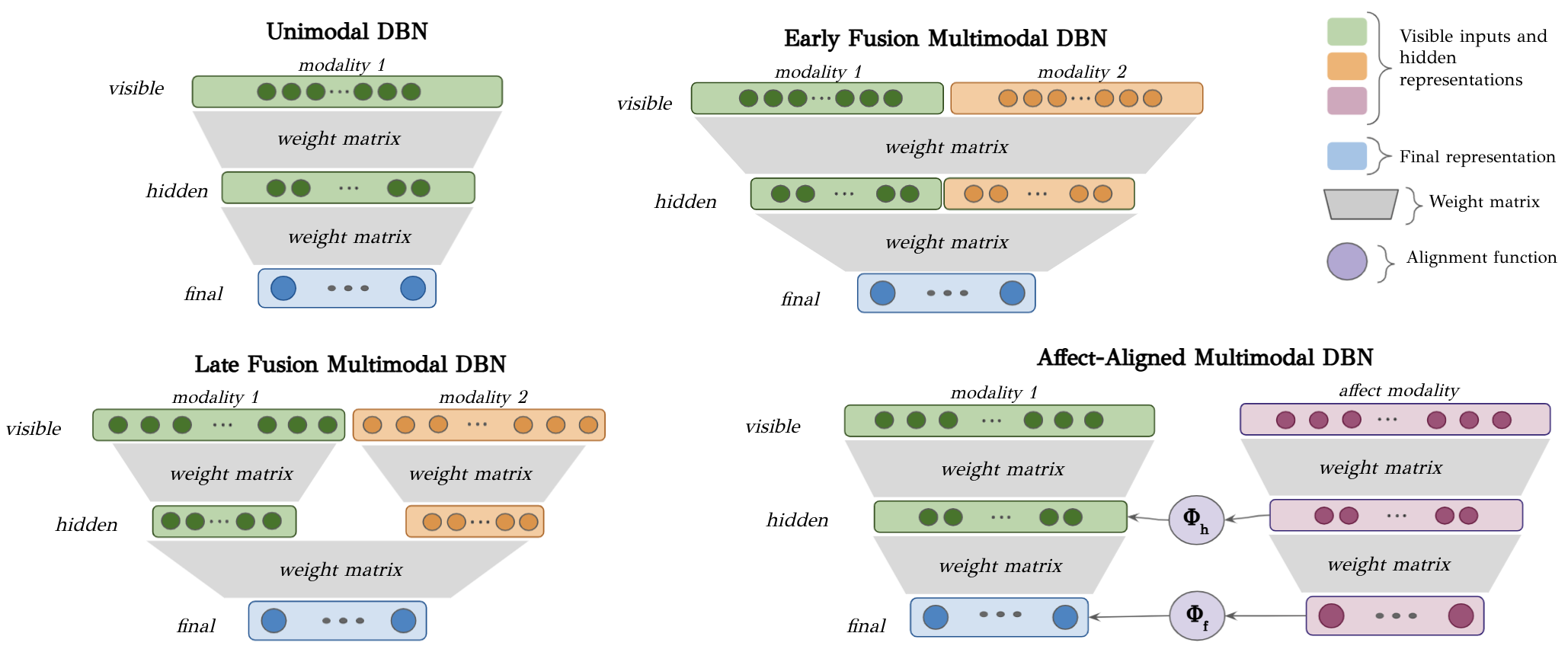}
   \caption{Visualization of Unimodal and Multimodal DBN Approaches.}
   \label{fig:models}
\end{figure*}
We experimented with affect-aware unimodal and multimodal DBN-based approaches for unsupervised high-stakes deception detection; models were trained on facial valence, facial arousal, audio, and visual features. In addition to using facial affect as a \textit{feature} on which models are trained, we also developed a multimodal DBN training procedure that uses facial affect as an \textit{aligner} of audio-visual representations.

\subsection{High-Stakes Deception Dataset}
We used the only publicly-available, real-world, high-stakes deception dataset of people speaking in 121 courtroom trial situations ($\sim$28 seconds  per video) discussed in Section II.A; additional details are in \cite{10.1145/2818346.2820758}. The creators of this dataset labeled each video as \textit{truthful} or \textit{deceptive}, based on information verified by police testimony and trials. We used their labels as ground truth to evaluate the effectiveness of our deception detection models. Videos were collected in unconstrained situations with a variety of illuminations, camera angles, camera movements, background settings, poses, and face occlusions. A usable subset, based on criteria defined by prior research with this dataset \cite{DBLP:conf/aaai/WuSDS18, 10.1145/3382507.3418864}, included 108 videos (53 truthful and 55 deceptive videos; 47 speakers of diverse race and gender).

\subsection{Multimodal Feature Extraction}
\subsubsection{Affect Features}
Similar to prior methods \cite{10.1145/3382507.3418864}, we implemented an AffWildNet model \cite{kollias2019deep} in TensorFlow \cite{tensorflow2015-whitepaper} trained on the Aff-Wild database to extract continuous representations of facial valence and facial arousal from speakers in the videos \cite{zafeiriou2017aff, kollias2017recognition}. The Aff-Wild data are labeled for continuous facial valence and facial arousal (between -1 and 1) at each facial frame, by 8 trained annotators. This data from situations in-the-wild included a variety of environments, poses, illuminations, and occlusions, and diverse speaker ethnicities. AffWildNet employs a CNN and RNN spatio-temporal architecture to predict valence and arousal at each facial frame. Tracking the faces of primary speakers, we extracted images of the facial bounding box at each video frame, resized images (96x96x3), and normalized pixel intensities between -1 and 1. Model weights came from the AffWildNet developers. We prepared Tensors with sequences of 80 consecutive images, per specified AffWildNet hyperparameters, and performed inference on these with AffWildNet to extract affect features. We extracted \textit{2} facial affect features (valence and arousal) at each video frame. 

\subsubsection{Audio Features}
We used the OpenSMILE toolkit \cite{10.1145/1873951.1874246} (version 2.0) to extract audio features from the eGeMAPs \cite{eyben2015geneva} and MFCC feature sets, guided by prior deception detection approaches \cite{RillGarca2019HighLevelFF, DBLP:conf/aaai/WuSDS18, 7836768, 10.1145/3382507.3418864}. These features capture cepstral, spectral, prosodic, energy, and voice quality information. OpenSMILE's eGeMAPs and MFCC feature sets overlap on MFCC features; we excluded duplicates. We extracted \textit{58} audio features at each audio frame. 

\subsubsection{Visual Features}
We used the OpenFace toolkit \cite{baltrusaitis2018openface} (version 2.2.0) to extract visual features that captured eye gaze, facial action units, and head pose information, guided by prior deception detection approaches \cite{RillGarca2019HighLevelFF, DBLP:conf/aaai/WuSDS18, 7836768, 10.1145/3382507.3418864, 10.1145/3349801.3349806}. We extracted \textit{31} visual features at each video frame. 

\subsubsection{Temporal Representations}
Similar to prior methods \cite{RillGarca2019HighLevelFF, 10.1145/3382507.3418864}, we chose to represent each affect, audio, and visual feature as a fixed-length vector of time-series attributes to capture their temporal behavior along each variable-length video. We computed 17 time-series attributes for each feature: mean, median, standard deviation, minimum, maximum, kurtosis, mean and median autocorrelation across 1-second time lags, and changes in feature values across quantiles. A total of 1547 multimodal features were computed to represent each video (17 facial valence, 17 facial arousal, 527 visual, and 986 audio features). 

\subsection{Multimodal DBN Representation Learning}
We extracted low-dimensional representations of each video by training DBNs with 4 modeling approaches: unimodal networks, multimodal early fusion networks, multimodal late fusion networks, and multimodal affect-aligned networks. Conceptual diagrams visualizing our approaches are in \textbf{Figure \ref{fig:models}} and described in this section. The goal of our DBN approaches is to capture complex, non-linear dependencies in visible, behavioral input data by learning lower dimensional, hidden representations. 

\subsubsection{DBN Training}
Each of our DBN approaches follow the same underlying training procedure. DBNs have stacks of RBMs that each have one visible layer \textbf{v} and one hidden layer \textbf{h}. The energy \cite{hopfield1982neural} of a joint configuration (\textbf{v}, \textbf{h}), given the weight matrix and bias parameters $\theta$ = (\textbf{W}, \textbf{a}, \textbf{b}), can be expressed with the following equation \cite{hinton2012practical, hinton2006fast}: 
\begin{equation} 
E(v,h;\theta) = -\sum_{i \in visible}a_{i}v_{i} -\sum_{j \in hidden}b_{j}h_{j}-\sum_{i,j}v_{i}h_{j}W_{ij}
\label{energy}
\end{equation}
This energy function can be used to express the joint probability of \textbf{v} and \textbf{h} as follows:
\begin{equation}
P(v,h;\theta) = \frac{1}{Z(\theta)}e^{-E(v,h;\theta)}
\label{prob}
\end{equation}
\begin{equation}
Z(\theta) = \sum_{v,h}e^{-E(v,h;\theta)}
\end{equation}
During training, we draw on \cite{hinton2012practical} to implement the contrastive divergence (CD) algorithm \cite{hinton2006fast}, approximating the gradient through Gibbs sampling. We use the stochastic gradient descent optimization algorithm to update the model's parameters. \textbf{Training details:} We trained DBNs with a greedy, layerwise approach, detailed in \cite{hinton2006fast, salakhutdinov2012efficient}. To allow comparisons of our DBN approaches, we kept the random seed (0) and hyperparameters constant during training (learning rate 0.01, 10 CD iterations, 200 epochs, batch size 32). 

\begin{figure*}[b]
   \centering
   \includegraphics[width=0.70\linewidth]{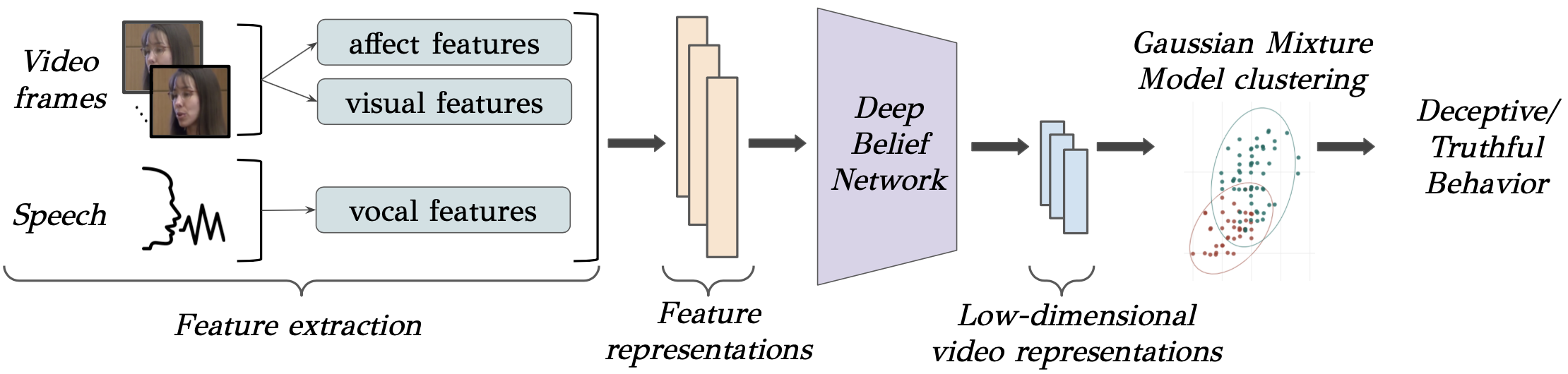}
   \caption{Process for Unsupervised Deception Detection.}
   \label{fig:process}
\end{figure*}

\subsubsection{DBN Architectures} To examine trends across modalities and explore the potential for  learning low-dimensional discriminative video representations, we report results from 2 and 4-dimensional embeddings from multiple DBN architectures. For unimodal, multimodal early fusion, and affect-aligned representations, we trained DBNs with 8 different architectures: 2 RBMs were trained to learn hidden representations of size \textit{2} and size \textit{4}; 6 DBNs were trained with stacked RBM architectures of 128-64-2, 256-128-2, 512-256-2, 128-64-4, 256-128-4, and  512-256-4. For multimodal late fusion representations, we trained 6 DBNs with the stacked RBM architectures, listed above. 
  
\subsubsection{Unimodal DBN} We trained unimodal DBN models on the respective input features of each modality to explore the effectiveness of learning representations of individual modalities for unsupervised deception detection.  

\subsubsection{Multimodal Early Fusion DBN} We experimented with early fusion DBNs which initially concatenate features from different modalities to form a single input on which DBNs are trained. We trained early fusion DBNs on each of the 11 combinations of the 4 modalities. Early fusion experiments were used to explore the effectiveness of exploiting low-level interactions of multimodal features to learn representations for unsupervised deception detection. 

\subsubsection{Multimodal Late Fusion DBN} We experimented with late fusion DBNs which train a separate unimodal RBM layer before concatenating the unimodal representations into a new input for joint DBN layers. Late fusion experiments were used to explore the effectiveness of exploiting higher-level interactions of multimodal feature spaces to learn representations for unsupervised deception detection. 

\subsubsection{Multimodal Affect-Aligned DBN} 
We formulated an approach for affect-aligned DBN representation learning by adapting the Kabsch algorithm \cite{kabsch1976solution} to learn audio, visual, and audio-visual DBN representations that are aligned with affect. For each affect-aligned approach, we concurrently train 2 DBN models: $DBN_{AV}$ is trained on audio, visual, or audio-visual input features and $DBN_{AFF}$ is trained on affect input features. During DBN training, we successively align the hidden representations of $DBN_{AV}$ with the corresponding layer of hidden representations in $DBN_{AFF}$. At each hidden layer \textit{$h_{1}...h_{n}$} of both DBNs, $DBN_{AV}$ will have a  \textit{D}-dimensional hidden representation $X = \left ( x^{i}\right )_{i=1}^{D}$, and $DBN_{AFF}$ will also have a \textit{D}-dimensional hidden representation $A = \left ( a^{i}\right )_{i=1}^{D}$. We compute an optimal alignment matrix $\phi$ to align $X$ with $A$ per the following: 
\begin{equation}
    \phi = argmin_{\phi}\left \| \phi X - A \right \|
\end{equation}
where $\left \| \cdot  \right \|$ denotes the Frobenius norm. We constrained this problem to consider rotation matrices as solutions, allowing us to use the Kabsch algorithm \cite{kabsch1976solution} to solve the equation. The covariance matrix $H$ of $X$ and $A$ is computed as follows:
\begin{equation}
    H  = [X - centroid(X)]^{T}[A - centroid(A)]
\end{equation}
The singular value decomposition of $H$ is then computed to express $H$ as a product of matrices $U$, $S$, and $V^{T}$. The optimal rotation matrix is then computed with the following: 
\begin{equation}
\phi = V\begin{bmatrix} 
    1 & 0 &\dots& 0\\
0 & 1 & \dots & 0\\
\vdots& \vdots & \ddots & \vdots\\
0&0&0&sign(det(VU^{T}))
    \end{bmatrix}U^T
\end{equation}
We use $\phi$ to align \textit{X} with \textit{A} during DBN training, resulting in DBN audio, visual, and audio-visual hidden representations that are successively aligned with affect at each DBN layer. Affect-alignment was efficiently implemented with NumPy and PyTorch \cite{NEURIPS2019_9015}.  We trained 9 affect-aligned DBNs that used (1) facial valence, (2) facial arousal, and (3) both facial valence and facial arousal to align audio, visual, and audio-visual modalities. These experiments were used to explore the effectiveness of affect as an \textit{aligner} of audio-visual representations for unsupervised deception detection.

\subsection{Unsupervised Classification Experiments}
To explore the extent to which our DBN representation learning approaches can be used for deception detection, we experimented with unsupervised GMMs \cite{hasselblad1966estimation} implemented in scikit-learn \cite{scikit-learn} to find clusters in representations. We compared the cluster assignments with the ground truth \textit{deceptive} and \textit{truthful} video labels to evaluate the performance of our models. Our process is visualized in \textbf{Figure 2}. 

GMMs have been widely used to evaluate representations during human-centered modeling tasks for speech \cite{5495662} and emotions \cite{4741177}, among other domains. Our clustering approach assumes that samples are generated through a mixture of two Gaussian distributions that are estimated with the Expectation-Maximization (EM) algorithm \cite{moon1996expectation}. We evaluate the quality of our representations by their effectiveness when clustering deceptive and truthful behavior; high-quality representations have a \textit{naturally clusterable} property \cite{bengio2013representation} (samples that come from the same class should be contained within distinguishable clusters in the representation space). \textbf{Training details:} We kept the random seed (1) and hyperparameters constant when training all GMM clustering models (diagonal covariance matrices, maximum of 100 EM iterations). Our methodology is similar to \cite{10.1145/3395035.3425191}, which experimented with unsupervised clustering algorithms for stress detection. 

Consistent with prior research \cite{8953413, 10.1145/3382507.3418864, RillGarca2019HighLevelFF, DBLP:conf/aaai/WuSDS18}, we ensured that the same person was not in both the training and testing sets of any experiment. All experiments were conducted with 5-fold stratified cross-validation, split across the 47 speaker identities, and repeated 10 times (50 cross-validation fold experiments). For feature normalization, training sets and testing sets for each cross-validation fold were scaled between 0 and 1 by the distribution of the training set, to prevent information from leaking into the testing sets. 

\subsection{Evaluation Metrics and Baselines}
Aligned with prior research \cite{10.1145/3382507.3418864, DBLP:conf/aaai/WuSDS18, RillGarca2019HighLevelFF}, we use \textit{AUC} as the primary metric to evaluate unsupervised deception detection classifiers:  the probability of the classifier ranking a randomly chosen deceptive sample higher than a randomly chosen truthful one. We computed this across 50 cross-validation folds; results are in \textbf{Table II}. Significance values of differences in model performances across representations were computed with McNemar's chi-squared test with continuity correction ($\alpha$=0.05) \cite{mcnemar}. We also compute  \textit{ACC}, the classification accuracy across the videos, and \textit{precision}, the proportion of deceptive videos among the videos classified as deceptive. Due to space limitation, we report accuracy and precision for the highest-performing approach. 

To compare the effectiveness of our DBN approaches to a simpler dimensionality-reduction approach, we defined a \textit{principal component analysis (PCA) baseline}, per prior methodology \cite{jaques2017multimodal}. We learned and clustered \textit{2} and \textit{4}-dimensional PCA  embeddings for each unimodal and multimodal combination of features, following the same experimental setup and cross-validation procedures. 

To compare our models to human deception detection ability at chance level, we defined a \textit{human performance baseline} as a classifier that always predicts \textit{deceptive} for 55 deceptive videos out of 108 videos (51\% accuracy) \cite{10.1145/3382507.3418864, 9413550}. It is worth noting that human accuracy on this dataset, averaged across audio, visual, and audio-visual modalities, has been determined as 52.78\% \cite{10.1145/2818346.2820758} from 3 human annotators and 70.07\% from a separate study with 3 different human annotators \cite{9165161}. The difference in these two measurements demonstrates the challenge of human deception detection. 

\section{RESULTS AND DISCUSSION}
\begin{table*}[h]
\centering
\caption{Classification Results (AUC)}
\begin{tabular}{l|c|c|c|c||cccc}
\textbf{Modality} & \multicolumn{4}{c||}{\textbf{Hidden Representation Size: 2}} & \multicolumn{4}{c}{\textbf{Hidden Representation Size: 4}} \\ \hline
 & \textbf{2 (RBM)} & \textbf{128\_64} & \textbf{256\_128} & \textbf{512\_256} & \multicolumn{1}{c|}{\textbf{4 (RBM)}} & \multicolumn{1}{c|}{\textbf{128\_64}} & \multicolumn{1}{c|}{\textbf{256\_128}} & \textbf{512\_256} \\ \hline
\multicolumn{9}{|c|}{\textbf{Unimodal Representations}} \\ \hline
\textbf{Arousal} & 0.47 & 0.47 & 0.50 & \cellcolor{UM}0.54 & \multicolumn{1}{c|}{0.51} & \multicolumn{1}{c|}{0.48} & \multicolumn{1}{c|}{0.53} & 0.46 \\
\textbf{Audio} & 0.46 & 0.43 & 0.45 & \cellcolor{UM}0.51 & \multicolumn{1}{c|}{0.48} & \multicolumn{1}{c|}{0.47} & \multicolumn{1}{c|}{0.41} & 0.48 \\
\textbf{Valence} & 0.68 & 0.65 & 0.67 & 0.60 & \multicolumn{1}{c|}{0.56} & \multicolumn{1}{c|}{\cellcolor{UM}0.69} & \multicolumn{1}{c|}{0.65} & 0.62 \\
\textbf{Visual} & 0.65 & 0.59 & 0.65 & 0.59 & \multicolumn{1}{c|}{\cellcolor{UM}0.70} & \multicolumn{1}{c|}{0.53} & \multicolumn{1}{c|}{0.61} & 0.61 \\ \hline
\multicolumn{9}{|c|}{\textbf{Multimodal Early Fusion Representations}} \\ \hline
\textbf{Arousal + Audio} & 0.50 & 0.50 & 0.45 & 0.49 & \multicolumn{1}{c|}{0.48} & \multicolumn{1}{c|}{0.48} & \multicolumn{1}{c|}{0.49} & 0.41 \\
\cellcolor{Best}\textbf{Arousal + Valence} & 0.50 & \cellcolor{Best} \textbf{0.69} & 0.61 & 0.74 & \multicolumn{1}{c|}{0.56} & \multicolumn{1}{c|}{\cellcolor{Best} \textbf{0.73}} & \multicolumn{1}{c|}{0.62} & 0.70 \\
\textbf{Arousal + Visual} & 0.56 & 0.64 & 0.70 & 0.65 & \multicolumn{1}{c|}{0.59} & \multicolumn{1}{c|}{0.57} & \multicolumn{1}{c|}{0.52} & 0.54 \\
\textbf{Audio + Valence} & 0.48 & 0.46 & 0.52 & 0.43 & \multicolumn{1}{c|}{0.48} & \multicolumn{1}{c|}{0.50} & \multicolumn{1}{c|}{0.46} & 0.45 \\
\textbf{Audio + Visual} & 0.52 & 0.50 & 0.47 & 0.54 & \multicolumn{1}{c|}{0.40} & \multicolumn{1}{c|}{0.50} & \multicolumn{1}{c|}{0.46} & 0.49 \\
\cellcolor{Best}\textbf{Valence + Visual} & 0.62 & 0.61 & 0.63 & 0.47 & \multicolumn{1}{c|}{\cellcolor{Best}\textbf{0.71}} & \multicolumn{1}{c|}{0.55} & \multicolumn{1}{c|}{0.52} & 0.60 \\
\textbf{Arousal + Audio + Valence} & 0.46 & 0.48 & 0.39 & 0.47 & \multicolumn{1}{c|}{0.48} & \multicolumn{1}{c|}{0.49} & \multicolumn{1}{c|}{0.53} & 0.45 \\
\textbf{Arousal + Audio + Visual} & 0.51 & 0.48 & 0.58 & 0.46 & \multicolumn{1}{c|}{0.44} & \multicolumn{1}{c|}{0.42} & \multicolumn{1}{c|}{0.51} & 0.49 \\
\textbf{Arousal + Valence + Visual} & 0.64 & 0.43 & 0.71 & 0.71 & \multicolumn{1}{c|}{0.69} & \multicolumn{1}{c|}{0.51} & \multicolumn{1}{c|}{0.61} & 0.62 \\
\textbf{Audio + Valence + Visual} & 0.52 & 0.47 & 0.52 & 0.46 & \multicolumn{1}{c|}{0.46} & \multicolumn{1}{c|}{0.44} & \multicolumn{1}{c|}{0.45} & 0.47 \\
\textbf{All Modalities} & 0.49 & 0.51 & 0.49 & 0.53 & \multicolumn{1}{c|}{0.40} & \multicolumn{1}{c|}{0.53} & \multicolumn{1}{c|}{0.48} & 0.55 \\ \hline
\multicolumn{9}{|c|}{\textbf{Multimodal Late Fusion Representations}} \\ \hline
\textbf{Arousal + Audio} & - & 0.46 & 0.52 & 0.50 & \multicolumn{1}{c|}{-} & \multicolumn{1}{c|}{0.38} & \multicolumn{1}{c|}{0.44} & 0.47 \\
\textbf{Arousal + Valence} & - & 0.58 & 0.73 & 0.49 & \multicolumn{1}{c|}{-} & \multicolumn{1}{c|}{0.50} & \multicolumn{1}{c|}{0.67} & 0.56 \\
\textbf{Arousal + Visual} & - & 0.56 & 0.71 & 0.76 & \multicolumn{1}{c|}{-} & \multicolumn{1}{c|}{0.54} & \multicolumn{1}{c|}{0.73} & 0.67 \\
\textbf{Audio + Valence} & - & 0.45 & 0.52 & 0.58 & \multicolumn{1}{c|}{-} & \multicolumn{1}{c|}{0.48} & \multicolumn{1}{c|}{0.45} & 0.50 \\
\textbf{Audio + Visual} & - & 0.47 & 0.47 & 0.59 & \multicolumn{1}{c|}{-} & \multicolumn{1}{c|}{0.49} & \multicolumn{1}{c|}{0.51} & 0.53 \\
\cellcolor{Best}\textbf{Valence + Visual} & - & 0.56 & 0.64 & \cellcolor{Best} \textbf{0.80*} & \multicolumn{1}{c|}{-} & \multicolumn{1}{c|}{0.52} & \multicolumn{1}{c|}{\cellcolor{Best}\textbf{0.75}} & 0.70 \\
\textbf{Arousal + Audio + Valence} & - & 0.50 & 0.49 & 0.50 & \multicolumn{1}{c|}{-} & \multicolumn{1}{c|}{0.46} & \multicolumn{1}{c|}{0.47} & 0.46 \\
\textbf{Arousal + Audio + Visual} & - & 0.45 & 0.63 & 0.63 & \multicolumn{1}{c|}{-} & \multicolumn{1}{c|}{0.59} & \multicolumn{1}{c|}{0.43} & 0.55 \\
\cellcolor{Best}\textbf{Arousal + Valence + Visual} & - & 0.62 & \cellcolor{Best} \textbf{0.79} & 0.62 & \multicolumn{1}{c|}{-} & \multicolumn{1}{c|}{0.62} & \multicolumn{1}{c|}{0.69} & \cellcolor{Best}\textbf{0.77} \\
\textbf{Audio + Valence + Visual} & - & 0.58 & 0.52 & 0.63 & \multicolumn{1}{c|}{-} & \multicolumn{1}{c|}{0.48} & \multicolumn{1}{c|}{0.58} & 0.53 \\
\textbf{All Modalities} & \multicolumn{1}{l|}{} & 0.50 & 0.49 & 0.50 & \multicolumn{1}{c|}{-} & \multicolumn{1}{c|}{0.46} & \multicolumn{1}{c|}{0.47} & 0.46 \\ \hline
\multicolumn{9}{|c|}{\textbf{Multimodal Affect-Aligned Representations}} \\ \hline
\textbf{Arousal + Audio} & 0.50 & 0.50 & 0.54 & 0.52 & \multicolumn{1}{c|}{0.48} & \multicolumn{1}{c|}{0.54} & \multicolumn{1}{c|}{0.51} & 0.51 \\
\cellcolor{Best}\textbf{Arousal + Visual} & \cellcolor{Best}\textbf{0.69} & 0.43 & 0.56 & 0.51 & \multicolumn{1}{c|}{0.70} & \multicolumn{1}{c|}{0.50} & \multicolumn{1}{c|}{0.38} & 0.52 \\
\textbf{Arousal + Audio-Visual} & 0.44 & 0.60 & 0.50 & 0.49 & \multicolumn{1}{c|}{0.56} & \multicolumn{1}{c|}{0.47} & \multicolumn{1}{c|}{0.52} & 0.52 \\
\textbf{Valence + Audio} & 0.50 & 0.50 & 0.56 & 0.55 & \multicolumn{1}{c|}{0.48} & \multicolumn{1}{c|}{0.46} & \multicolumn{1}{c|}{0.44} & 0.54 \\
\cellcolor{Best}\textbf{Valence + Visual} & 0.62 & 0.48 & 0.48 & 0.35 & \multicolumn{1}{c|}{0.66} & \multicolumn{1}{c|}{0.51} & \multicolumn{1}{c|}{\cellcolor{Best} \textbf{0.75}} & 0.45 \\
\textbf{Valence + Audio-Visual} & 0.43 & 0.50 & 0.50 & 0.60 & \multicolumn{1}{c|}{0.47} & \multicolumn{1}{c|}{0.51} & \multicolumn{1}{c|}{0.56} & 0.57 \\
\textbf{Arousal \& Valence + Audio} & 0.45 & 0.47 & 0.46 & 0.55 & \multicolumn{1}{c|}{0.48} & \multicolumn{1}{c|}{0.44} & \multicolumn{1}{c|}{0.48} & 0.44 \\
\textbf{Arousal \& Valence + Visual} & 0.67 & 0.62 & 0.54 & 0.56 & \multicolumn{1}{c|}{0.59} & \multicolumn{1}{c|}{0.64} & \multicolumn{1}{c|}{0.53} & 0.55 \\
\textbf{Arousal \& Valence + Audio-Visual} & 0.44 & 0.59 & 0.62 & 0.62 & \multicolumn{1}{c|}{0.41} & \multicolumn{1}{c|}{0.47} & \multicolumn{1}{c|}{0.45} & 0.44
\end{tabular}
\end{table*}
AUC results from the classification experiments are presented in \textbf{Table II}. The highest-performing representation for each DBN  architecture (column) is highlighted in orange. \textit{Across all DBN architectures, the highest-performing representations always included facial affect, supporting the inclusion of affect as a modality when learning discriminative DBN representations for unsupervised high-stakes deception detection}. Our best representations were learned by DBNs with a \textit{512$-$256$-$2} architecture trained with multimodal late fusion on facial valence and visual features. Unsupervised GMM classifiers trained on these representations achieved an AUC of 80\% (accuracy of 70\% and precision of 88\%), outperforming the 51\% human performance baseline and the corresponding PCA baseline of 54\% ($p<0.01$). \textit{Our results demonstrate the potential for DBN-based models to effectively learn representations of deceptive and truthful behavior for unsupervised deception detection.}

\subsection{Analysis of Unimodal Representations}
Unimodal classification results revealed that models trained on facial valence and visual representations significantly outperformed models trained on facial arousal and audio representations across DBN architectures ($p<0.001$). Differences in performance are visualized in \textbf{Figure 3}. The highest AUCs achieved by classifiers trained on unimodal visual, facial valence, facial arousal, and audio representations were 70\%, 69\%, 54\%, and 51\%, respectively (highlighted in blue in \textbf{Table II}). Visualizations of these classifier performances compared to their corresponding PCA baselines are also in \textbf{Figure 3}. Classifiers trained on the highest-performing visual, facial valence, and facial arousal representations significantly outperformed their corresponding PCA baselines of 49\%, 49\%, and 43\%, respectively ($p<0.001$, $p<0.001$, $p<0.05$). \textit{Our results suggest that unimodal DBN models can effectively learn discriminative representations of deceptive and truthful behavior. Results indicate that facial affect contributes to the quality of DBN representations when used as a feature for high-stakes deception detection.}

\begin{figure}[h]
   \centering
   \includegraphics[width=0.70\linewidth]{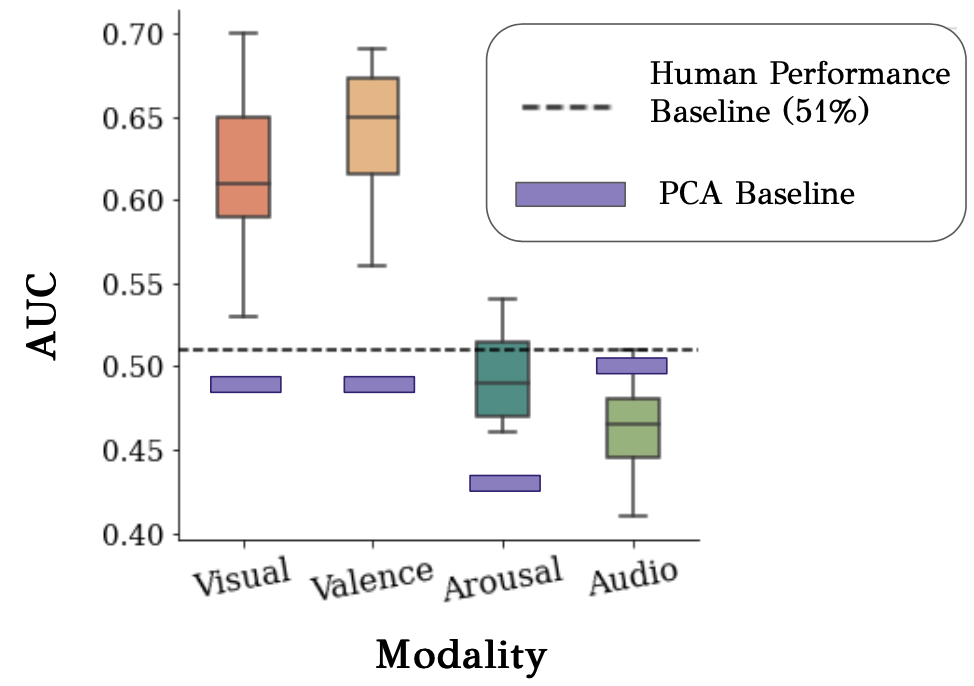}
   \caption{Distribution of AUC performances achieved by unimodal representations across all RBM/DBM architectures.}
   \label{fig:umodal}
\end{figure}

Classifiers trained on the highest-performing audio representations did not significantly outperform the corresponding PCA baseline of 50\%, indicating that audio representations were not as discriminative for detecting high-stakes deception in this dataset's  context (courtroom trials). It is worth noting that audio cues have been useful for detecting deception in other contexts \cite{DBLP:conf/aaai/WuSDS18, 10.1145/3382507.3418864, RillGarca2019HighLevelFF, 7836768}. 

Classifiers trained on facial valence representations outperformed classifiers trained on visual representations across all DBN architectures, with the exception of the 4-dimensional visual RBM representation. Comparable performances of facial valence and visual representations suggest that they have comparable discriminative potential for use in unsupervised deception detection.

\textit{Our findings from these  unimodal classification experiments support the inclusion of affect, specifically facial valence, as a feature to be used in DBN-based representation learning for high-stakes deception detection in the wild.} 

\subsection{Analysis of Multimodal Representations}
Across the 8 DBN architectures, multimodal representations outperformed all unimodal representations, demonstrating that multimodal models are more capable of learning useful representations of human behavior than unimodal ones \cite{mmml}. Four modality combinations, highlighted in orange, emerged as most effective: (valence + visual), (arousal + valence + visual), (arousal + valence), (arousal + visual). The highest-performing multimodal representations significantly outperformed their corresponding PCA baselines ($p<0.01$), visualized in \textbf{Figure 4}. In all of the highest-performing representations across each DBN architecture, affect was either a \textit{feature} or an \textit{aligner} within the modality combination. 

\begin{figure}[h]
   \centering
   \includegraphics[width=0.99\linewidth]{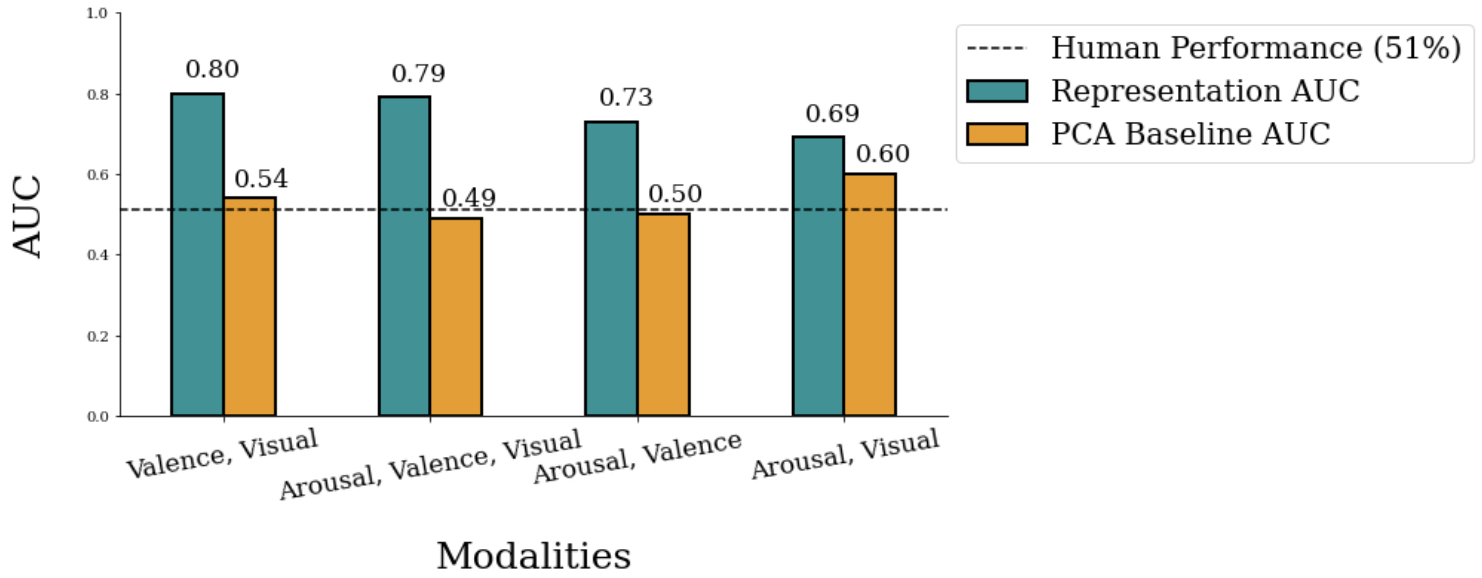}
   \caption{AUC performance achieved by highest-performing multimodal DBN representations relative to their baselines.}
   \label{fig:umodal}
\end{figure}

We examined the performance of our models from the 3 multimodal representation learning approaches: early fusion, late fusion, and affect-aligned representations. Across the 8 DBN architectures, early fusion  yielded the highest performance 3 times, late fusion yielded the highest performance 4 times, and affect-aligned representations yielded the highest performance 2 times (there was one tie).

\break 
\textit{Our findings provide support for the inclusion of affect as a feature or aligner in multimodal machine learning approaches to learn useful representations of human behavior}. 

Comparing our results to those obtained from prior fully-supervised high-stakes deception detection approaches (Table I), we demonstrate the potential for modeling high-stakes deception with unsupervised, affect-aware DBN representation learning. While some prior fully-supervised models outperform our unsupervised models, our highest-performing AUC of 80\% falls  within the range achieved by the highest-performing prior results (70\%$-$91\%). All these prior models required labeled datasets for training; \textit{our approach has the advantage that it does not require any labeled data.} 

\section{CONCLUSION}
Our paper addresses the scarcity of labeled, real-world high-stakes deception data for training models, by developing the first unsupervised approaches that detect high-stakes deception without requiring labels. We experimented with unimodal and multimodal DBN-based approaches that were trained on facial valence, facial arousal, audio, and visual features to identify effective modeling approaches for unsupervised deception detection. In addition to using facial affect as a \textit{feature} on which DBN models are trained, we also introduced a multimodal DBN training procedure that used facial affect as an \textit{aligner} of audio-visual representations. We demonstrated that unsupervised DBN approaches can learn discriminative, low-dimensional representations of deceptive and truthful behavior. Our best unsupervised approach achieved an AUC of 80\%, accuracy of 70\%, and precision of 88\% (trained on facial valence and visual features), outperforming human ability and performing comparably to a number of existing fully-supervised high-stakes deception detection models. To the best of our knowledge, our research also represents the first attempt at training DBNs to learn discriminative representations of a social behavior. 

In all of the highest-performing approaches across each DBN architecture, affect emerged as a contributor, functioning either as a feature or as an aligner, supporting the inclusion of facial affect as a modality when learning discriminative DBN representations for unsupervised high-stakes deception detection. While this paper used facial affect, rich affect features may also be gained from the audio modality, motivating future work in this direction.  

Our findings serve as a proof-of-concept and motivation for future research in developing unsupervised, affect-aware machine learning approaches for representing and detecting deception and other social behaviors during interaction contexts in real-world situations. We are also encouraged by the findings to proceed beyond our paper's research context to examine the role of affect as a feature and aligner in models of human behavior, beyond deception, in-the-wild. 
\section{ACKNOWLEDGMENTS}
USC Provost's Undergraduate Research Fellowship.


{\small
\bibliographystyle{ieee}
\bibliography{references}

\begin{thebibliography}{10}\itemsep=-1pt

\bibitem{tensorflow2015-whitepaper}
M.~Abadi et~al.
\newblock {TensorFlow}: Large-scale machine learning on heterogeneous systems,
  2015.
\newblock Software available from tensorflow.org.

\bibitem{Amiriparian+2016}
S.~Amiriparian, J.~Pohjalainen, E.~Marchi, S.~Pugachevskiy, and B.~Schuller.
\newblock Is deception emotional? an emotion-driven predictive approach.
\newblock In {\em Interspeech 2016}, pages 2011--2015, 2016.

\bibitem{App2011NonverbalCU}
B.~App, D.~McIntosh, C.~Reed, and M.~J. Hertenstein.
\newblock Nonverbal channel use in communication of emotion: how may depend on
  why.
\newblock {\em Emotion}, 11 3:603--17, 2011.

\bibitem{durante}
V.~{Ardulov}, Z.~{Durante}, S.~{Williams}, T.~{Lyon}, and S.~{Narayanan}.
\newblock Identifying truthful language in child interviews.
\newblock In {\em IEEE Int. Conf. on Acoust., Speech and Signal Process.},
  2020.

\bibitem{10.1145/3349801.3349806}
D.~{Avola} et~al.
\newblock Automatic deception detection in rgb videos using facial action
  units.
\newblock In {\em ACM Int. Conf. on Distrib. Smart Cameras}, 2019.

\bibitem{baltrusaitis2018openface}
T.~Baltrusaitis, A.~Zadeh, Y.~C. Lim, and L.-P. Morency.
\newblock Openface 2.0: Facial behavior analysis toolkit.
\newblock In {\em IEEE Int. Conf. on Autom. Face \& Gesture Recognition}, pages
  59--66. IEEE, 2018.

\bibitem{mmml}
T.~Baltrušaitis, C.~Ahuja, and L.-P. Morency.
\newblock Multimodal machine learning: A survey and taxonomy, 2017.

\bibitem{bengio2013representation}
Y.~Bengio, A.~Courville, and P.~Vincent.
\newblock Representation learning: A review and new perspectives.
\newblock {\em IEEE Trans. on Pattern Anal. and Mach. Intell.},
  35(8):1798--1828, 2013.

\bibitem{Bond2006AccuracyOD}
C.~{Bond} and B.~{Depaulo}.
\newblock Accuracy of deception judgments.
\newblock {\em Personality and Social Psychol. Rev.}, 10(3), 214-34, 2006.

\bibitem{10.1145/3107990.3108005}
M.~Burzo, M.~Abouelenien, V.~Perez-Rosas, and R.~Mihalcea.
\newblock {\em Multimodal Deception Detection}, page 419–453.
\newblock 2018.

\bibitem{8953413}
M.~{Ding}, A.~{Zhao}, Z.~{Lu}, T.~{Xiang}, and J.~{Wen}.
\newblock Face-focused cross-stream network for deception detection in videos.
\newblock In {\em IEEE Conf. on Comput. Vision and Pattern Recognition}, pages
  7794--7803, 2019.

\bibitem{mcnemar}
W.~{Dupont} and W.~{Plummer}.
\newblock Power and sample size calculations. a rev. and computer program.
\newblock {\em Controlled Clin. Trials}, 11:116--28, 1990.

\bibitem{ekman1969nonverbal}
P.~Ekman and W.~V. Friesen.
\newblock Nonverbal leakage and clues to deception.
\newblock {\em Psychiatry}, 32(1):88--106, 1969.

\bibitem{eyben2015geneva}
F.~Eyben et~al.
\newblock The geneva minimalistic acoustic parameter set (gemaps) for voice
  research and affective computing.
\newblock {\em IEEE Trans. on Affect. Comput.}, 7(2):190--202, 2015.

\bibitem{10.1145/1873951.1874246}
F.~Eyben, M.~W\"{o}llmer, and B.~Schuller.
\newblock Opensmile: The munich versatile and fast open-source audio feature
  extractor.
\newblock In {\em ACM Int. Conf. on Multimedia}, page 1459–1462, 2010.

\bibitem{forgas1995mood}
J.~P. Forgas.
\newblock Mood and judgment: the affect infusion model (aim).
\newblock {\em Psychological Bull.}, 117(1):39, 1995.

\bibitem{ekmanhs}
M.~Frank and P.~Ekman.
\newblock The ability to detect deceit generalizes across different types of
  high-stake lies.
\newblock {\em Personality and Social Psychol.}, 72:1429--39, 07 1997.

\bibitem{glavas-vulic-2020-non}
G.~Glava{\v{s}} and I.~Vuli{\'c}.
\newblock Non-linear instance-based cross-lingual mapping for non-isomorphic
  embedding spaces.
\newblock In {\em Annu. Meeting of the Assoc. for Comput. Linguistics}, pages
  7548--7555, 2020.

\bibitem{hasselblad1966estimation}
V.~Hasselblad.
\newblock Estimation of parameters for a mixture of normal distributions.
\newblock {\em Technometrics}, 8(3):431--444, 1966.

\bibitem{hinton2012practical}
G.~E. Hinton.
\newblock A practical guide to training restricted boltzmann machines.
\newblock In {\em Neural networks: Tricks of the trade}. Springer, 2012.

\bibitem{hinton2006fast}
G.~E. Hinton, S.~Osindero, and Y.-W. Teh.
\newblock A fast learning algorithm for deep belief nets.
\newblock {\em Neural Comput.}, 18(7):1527--1554, 2006.

\bibitem{hopfield1982neural}
J.~J. Hopfield.
\newblock Neural networks and physical symp. with emergent collective
  computational abilities.
\newblock {\em Proc. of the Nat. Acad. of Sciences}, 79(8):2554--2558, 1982.

\bibitem{6639140}
J.~Huang and B.~Kingsbury.
\newblock Audio-visual deep learning for noise robust speech recognition.
\newblock In {\em IEEE Int. Conf. on Acoust., Speech and Signal Process.},
  pages 7596--7599, 2013.

\bibitem{7836768}
M.~{Jaiswal}, S.~{Tabibu}, and R.~{Bajpai}.
\newblock The truth and nothing but the truth: Multimodal analysis for
  deception detection.
\newblock In {\em IEEE Int. Conf. on Data Mining Workshops}, pages 938--943,
  2016.

\bibitem{jaques2017multimodal}
N.~Jaques, S.~Taylor, A.~Sano, and R.~Picard.
\newblock Multimodal autoencoder: A deep learning approach to filling in
  missing sensor data and enabling better mood prediction.
\newblock In {\em Int. Conf. on Affect. Comput. and Intell. Interaction}, pages
  202--208. IEEE, 2017.

\bibitem{kabsch1976solution}
W.~Kabsch.
\newblock A solution for the best rotation to relate two sets of vectors.
\newblock {\em Acta Crystallographica Sect. A: Crystal Physics, Diffraction,
  Theor. and General Crystallography}, 32(5):922--923, 1976.

\bibitem{8621909}
H.~Karimi, J.~Tang, and Y.~Li.
\newblock Toward end-to-end deception detection in videos.
\newblock In {\em IEEE Int. Conf. on Big Data}, pages 1278--1283, 2018.

\bibitem{Kheder2016}
W.~B. Kheder, D.~Matrouf, M.~Ajili, and J.-F. Bonastre.
\newblock Iterative bayesian and mmse-based noise compensation techniques for
  speaker recognition in the i-vector space.
\newblock In {\em Odyssey 2016}, pages 60--67, 2016.

\bibitem{6638346}
Y.~Kim, H.~Lee, and E.~M. Provost.
\newblock Deep learning for robust feature generation in audiovisual emotion
  recognition.
\newblock In {\em IEEE Int. Conf. on Acoust., Speech and Signal Process.},
  pages 3687--3691, 2013.

\bibitem{kollias2019deep}
D.~Kollias et~al.
\newblock Deep affect prediction in-the-wild: Aff-wild database and challenge,
  deep architectures, and beyond.
\newblock {\em Int. J. of Comput. Vision}, pages 1--23, 2019.

\bibitem{kollias2017recognition}
D.~Kollias, M.~A. Nicolaou, I.~Kotsia, G.~Zhao, and S.~Zafeiriou.
\newblock Recognition of affect in the wild using deep neural networks.
\newblock In {\em IEEE Conf. on Comput. Vision and Pattern Recognition
  Workshops}, pages 1972--1979. IEEE, 2017.

\bibitem{10.1145/3382507.3418864}
L.~Mathur and M.~J. Matari\'{c}.
\newblock Introducing representations of facial affect in automated multimodal
  deception detection.
\newblock In {\em Int. Conf. on Multimodal Interaction}, page 305–314, 2020.

\bibitem{9413550}
L.~Mathur and M.~J. Matarić.
\newblock Unsupervised audio-visual subspace alignment for high-stakes
  deception detection.
\newblock In {\em IEEE Int. Conf. on Acoust., Speech and Signal Process.},
  pages 2255--2259, 2021.

\bibitem{4741177}
A.~Metallinou, S.~Lee, and S.~Narayanan.
\newblock Audio-visual emotion recognition using gaussian mixture models for
  face and voice.
\newblock In {\em IEEE Int. Symp. on Multimedia}, pages 250--257, 2008.

\bibitem{moon1996expectation}
T.~K. Moon.
\newblock The expectation-maximization algorithm.
\newblock {\em IEEE Signal Process. Mag.}, 13(6):47--60, 1996.

\bibitem{10.1109/CVPR.2014.299}
W.~Ouyang, X.~Chu, and X.~Wang.
\newblock Multi-source deep learning for human pose estimation.
\newblock In {\em IEEE Conf. on Comput. Vision and Pattern Recognition}, page
  2337–2344, USA, 2014.

\bibitem{NEURIPS2019_9015}
A.~{Paszke} et~al.
\newblock Pytorch: An imperative style, high-performance deep learning library.
\newblock In {\em Advances in Neural Inf. Process. Symp. 32}, pages 8024--8035.
  2019.

\bibitem{scikit-learn}
F.~Pedregosa et~al.
\newblock Scikit-learn: Machine learning in {P}ython.
\newblock {\em J. of Mach. Learn. Res.}, 12:2825--2830, 2011.

\bibitem{10.1145/2818346.2820758}
V.~{P\'{e}rez-Rosas} et~al.
\newblock Deception detection using real-life trial data.
\newblock In {\em ACM Int. Conf. on Multimodal Interaction}, page 59–66,
  2015.

\bibitem{picard2000affective}
R.~W. Picard.
\newblock {\em Affective Computing}.
\newblock MIT press, 2000.

\bibitem{porter_brinke}
S.~{Porter} and L.~{ten Brinke}.
\newblock The truth about lies: What works in detecting high-stakes deception?
\newblock {\em Legal and Criminological Psych.}, 15(1):57--75, 2010.

\bibitem{5495662}
D.~Povey et~al.
\newblock Subspace gaussian mixture models for speech recognition.
\newblock In {\em IEEE Int. Conf. on Acoust., Speech and Signal Process.},
  pages 4330--4333, 2010.

\bibitem{RillGarca2019HighLevelFF}
R.~{Rill-Garc{\'i}a} et~al.
\newblock High-level features for multimodal deception detection in videos.
\newblock {\em IEEE Conf. on Comput. Vision and Pattern Recognition Workshops},
  pages 1565--1573, 2019.

\bibitem{russell1980circumplex}
J.~A. Russell.
\newblock A circumplex model of affect.
\newblock {\em Personality and Social Psychol.}, 39(6):1161, 1980.

\bibitem{salakhutdinov2015learning}
R.~Salakhutdinov.
\newblock Learning deep generative models.
\newblock {\em Annu. Rev. of Statist. and Its Application}, 2:361--385, 2015.

\bibitem{salakhutdinov2009deep}
R.~Salakhutdinov and G.~Hinton.
\newblock Deep boltzmann machines.
\newblock In {\em Artif. Intell. and Statist.}, pages 448--455. PMLR, 2009.

\bibitem{salakhutdinov2012efficient}
R.~Salakhutdinov and G.~Hinton.
\newblock An efficient learning procedure for deep boltzmann machines.
\newblock {\em Neural Comput.}, 24(8):1967--2006, 2012.

\bibitem{9165161}
U.~M. Sen, V.~Perez-Rosas, B.~Yanikoglu, M.~Abouelenien, M.~Burzo, and
  R.~Mihalcea.
\newblock Multimodal deception detection using real-life trial data.
\newblock {\em IEEE Trans. on Affect. Comput.}, 2020.

\bibitem{srivastava2012multimodal}
N.~Srivastava and R.~Salakhutdinov.
\newblock Multimodal learning with deep boltzmann machines.
\newblock In {\em NIPS}, volume~1, page~2. Citeseer, 2012.

\bibitem{taylor2012vitruvian}
J.~Taylor, J.~Shotton, T.~Sharp, and A.~Fitzgibbon.
\newblock The vitruvian manifold: Inferring dense correspondences for one-shot
  human pose estimation.
\newblock In {\em IEEE Conf. on Comput. Vision and Pattern Recognition}, pages
  103--110. IEEE, 2012.

\bibitem{10.1145/1459359.1459573}
A.~Vinciarelli, M.~Pantic, H.~Bourlard, and A.~Pentland.
\newblock Social signal processing: State-of-the-art and future perspectives of
  an emerging domain.
\newblock In {\em ACM Int. Conf. on Multimedia}, page 1061–1070, 2008.

\bibitem{vrij2006police}
A.~Vrij, L.~Akehurst, and S.~Knight.
\newblock Police officers', social workers', teachers' and the general public's
  beliefs about deception in children, adolescents and adults.
\newblock {\em Legal and Criminological Psychol.}, 11(2):297--312, 2006.

\bibitem{10.1145/2647868.2654969}
D.~Wu and L.~Shao.
\newblock Multimodal dynamic networks for gesture recognition.
\newblock In {\em ACM Int. Conf. on Multimedia}, page 945–948, 2014.

\bibitem{10.1145/3395035.3425191}
Y.~Wu, M.~Daoudi, A.~Amad, L.~Sparrow, and F.~D'Hondt.
\newblock Unsupervised learning method for exploring students' mental stress in
  medical simulation training.
\newblock In {\em Companion Publication of the 2020 Int. Conf. on Multimodal
  Interaction}, page 165–170, 2020.

\bibitem{DBLP:conf/aaai/WuSDS18}
Z.~Wu et~al.
\newblock Deception detection in videos.
\newblock In {\em AAAI Conf. on Artif. Intell.}, pages 1695--1702, 2018.

\bibitem{zafeiriou2017aff}
S.~Zafeiriou, D.~Kollias, M.~A. Nicolaou, A.~Papaioannou, G.~Zhao, and
  I.~Kotsia.
\newblock Aff-wild: Valence and arousal ‘in-the-wild’challenge.
\newblock In {\em IEEE Conf. on Comput. Vision and Pattern Recognition
  Workshops}, pages 1980--1987. IEEE, 2017.

\bibitem{zuckerman1981verbal}
M.~Zuckerman, B.~M. DePaulo, and R.~Rosenthal.
\newblock Verbal and nonverbal communication of deception.
\newblock In {\em Advances in Exp. Social Psychol.}, volume~14, pages 1--59.
  Elsevier, 1981.

\end{thebibliography}
}
\end{document}